\documentclass[letterpaper]{article} 
\usepackage{aaai23}  
\usepackage{times}  
\usepackage{helvet}  
\usepackage{courier}  
\usepackage[hyphens]{url}  
\usepackage{graphicx} 
\usepackage{natbib}  
\usepackage{caption} 
\usepackage{comment}
\usepackage{amsmath}
\usepackage{amsfonts}
\usepackage{amssymb}
\usepackage{amsbsy}
\usepackage{xcolor}
\usepackage{mathtools}
\usepackage{multirow}
\usepackage{stackengine}
\def\delequal{\mathrel{\ensurestackMath{\stackon[1pt]{=}{\scriptstyle\Delta}}}}
\usepackage{algorithm}
\usepackage{algorithmic}

\usepackage{newfloat}
\usepackage{listings}
\DeclareCaptionStyle{ruled}{labelfont=normalfont,labelsep=colon,strut=off} 
\floatstyle{ruled}
\newfloat{listing}{tb}{lst}{}
\floatname{listing}{Listing}
\title{Hybrid hidden Markov LSTM for short-term traffic flow prediction}
\author{
    Agnimitra Sengupta\textsuperscript{\rm 1}, Adway Das\textsuperscript{\rm 2}\equalcontrib, S. Ilgin Guler\textsuperscript{\rm 3}\equalcontrib\\
}
\affiliations{
    \textsuperscript{\rm 1,2,3}Department of Civil \& Environmental Engineering, The Pennsylvania State University\\
    University Park, PA 16802, USA\\
    \textsuperscript{\rm 1}avs6897@psu.edu, \textsuperscript{\rm 2}amd7293@psu.edu, \textsuperscript{\rm 3}sig123@psu.edu
}

\usepackage{bibentry}

\begin{document}

\maketitle

\begin{abstract}
Deep learning (DL) methods have outperformed parametric models such as historical average, ARIMA and variants in predicting traffic variables into short and near-short future, that are critical for traffic management. Specifically, recurrent neural network (RNN) and its variants (e.g. long short-term memory) are designed to retain long-term temporal correlations and therefore are suitable for modeling sequences. However, multi-regime models assume the traffic system to evolve through multiple states (say, free-flow, congestion in traffic) with distinct characteristics, and hence, separate models are trained to characterize the traffic dynamics within each regime. For instance, Markov-switching models with a hidden Markov model (HMM) for regime identification is capable of capturing complex dynamic patterns and non-stationarity. Interestingly, both HMM and LSTM can be used for modeling an observation sequence from a set of latent or, hidden state variables. In LSTM, the latent variable is computed in a deterministic manner from the current observation and the previous latent variable, while, in HMM, the set of latent variables is a Markov chain. Inspired by research in natural language processing, a hybrid hidden Markov-LSTM model that is capable of learning complementary features in traffic data is proposed for traffic flow prediction. Results indicate significant performance gains in using hybrid architecture compared to conventional methods such as Markov switching ARIMA and LSTM. 

\end{abstract}

\section{Introduction}

Accurate traffic predictions in the short or near-short term future, spanning from 5 minutes to 1 hour, play a vital role in efficient traffic management, encompassing traffic control and congestion mitigation. The effectiveness of various traffic control strategies, such as ramp metering or detour suggestions, heavily depends on precise traffic forecasting in the near future. However, achieving precise forecasting across both free-flow and congested traffic states is often challenging due to the inherent uncertainty and chaotic characteristics of transportation systems.
Numerous statistical models, both parametric and non-parametric, have been developed to accurately model the temporal aspects of traffic data. 
Parametric models including historical average (HA) algorithms, autoregressive integrated moving average (ARIMA) \cite{ahmed1979,levin1980} fails to uncover complex traffic dynamics as shown in \cite{davis1991,hamed1995}. To partially adapt to the complexities of traffic dynamics, multi-variable prediction models \cite{innamaa2000, dougherty1997, florio1996, lyons1996} and state-space models \cite{okutani1984, chen2001, chien2003} were developed. Additionally, trend retrieval using simple-average, principle component analysis (PCA), and wavelet methods have been discussed in the literature \cite{Chen2012, xie2007} to account for the apparent similarity of daily traffic flow time series. 

Alternately, multi-regime prediction models assume that a traffic system evolves through multiple regimes or states (say, free-flow, congestion) with distinct characteristics, and separate regression models are developed to predict traffic flow within each regime \cite{cetin2006, Kamarianakis2010,kamarianakis2012}. These model often use a Hidden Markov model (HMM) \cite{Rabiner1989} for the identification of traffic regimes. For example, see \cite{li2014,qi2014,zhu2016}. Although the multi-regime models are observed to identify the local trend within the time series more efficiently, the overall performance of these models was not significantly improved due to errors incurred while switching between regimes \cite{li2014}. 

Despite their reasonable performances, the specific functional form and methodological assumptions of the parametric models limit their capabilities to adapt to non-linearities associated with short-term trends -- which is a major shortcoming. Moreover, traffic data is observed to exhibit chaotic behaviour during congestion which makes it highly unstable \cite{disbro1989}. 
To the contrary, non-parametric techniques do not specify any functional form, rather they rely on pattern recognition to handle large data quantities. As a result, these approaches can better model traffic patterns with greater transferability and robustness across datasets \cite{smith1997, clark2003}. Nearest neighbors algorithms \cite{smith2002}, support vector machine \cite{mingheng2013}, and Bayesian network \cite{sun2006} are among the machine learning (ML) models that have demonstrated successful performances in small-scale traffic prediction problems. However, the success of such models often depend on suitable feature definition that require engineering judgement \cite{lin2019}.
Deep learning (DL) models use a multi-layer neural framework to capture complex relations in non-linear data \cite{lecun2015}, that require few to negligible feature engineering. Following its success, DL models have been extensively used in traffic time series modeling \cite{chang1995,innamaa2000,dia2001,park1999}. Specifically, recurrent neural networks (RNN) \cite{rumelhart1986} and its variants like long short-term memory (LSTM) \cite{hochreiter1997} are designed to preserve temporal correlations between observations in a time series, and hence better suited for traffic forecasting as well \cite{ma2015,yu2017, cui2020,yao2019,yao2018}.


Recent researches in natural language processing highlights the structural similarity between RNN and HMM, and their capability to learn complementary features from language data \cite{krakovna2016,Achille2019}. TConsequently, the use of hybrid models combining both RNN (specifically, Long Short-Term Memory or LSTM) and HMM can provide improved modeling capabilities for complex sequential data, such as traffic time series. 
This study focuses on investigating hybrid models that leverage the joint usage of HMM and LSTM for the task of traffic flow prediction. Additionally, we explore the feasibility of incorporating duration-based state transitions within the HMM framework in these models. 

The remainder of the paper is organized as follows. First, we present an overview on hidden (semi-) Markov models and LSTM, followed by the proposed hybrid models in this study. Next, the performance of hybrid models are compared with the baseline models. Finally, some concluding remarks are presented. 

\section{Background}
In this section, we provide a background on hidden Markov models (HMMs) and discuss the sojourn time distributions that are considered in this study. Subsequently, we present an overview of the long short-term memory (LSTM) model, which serves as the basis for the modifications that will be described in the following section.

\subsection{Hidden Markov models} \label{section:markov}

Markov chains model dynamical systems with the assumption that the state of the system at a time $t$ \textit{only} depends on the state in the immediately prior time step, $t-1$. However, such an assumption often does not hold true for complex dynamic systems. An alternative to Markov chains, the hidden Markov model (HMM) \cite{Rabiner1989, Rabiner1993} assumes the existence of a latent (hidden) process that follows a Markov chain from which observations $\mathbf{X}$ are generated. Therefore, for an observation sequence $\mathbf{X}=\{x_1,x_2,\cdots,x_T\}$ in $[1,T]$, there exists an unobserved state sequence $\mathbf{Z}=\{z_1,z_2,\cdots,z_T\}$, where the hidden states, $z_t$ belonging to state-space $Q\delequal\{q_1,q_2,\cdots,q_M\}$ follow a Markov chain governed by:
\begin{itemize}
    \item a state-transition probability matrix $\mathbf{A}=[a_{ij}]\in \mathbb{R}^{M\times M}$ where $a_{ij}= p(z_{t+1}=q_j\mid z_t=q_i)$
    \item initial state matrix $\pi=[\pi_i]\in\mathbb{R}^{1\times M}$ with $\pi_i=p(z_1=q_i)$ (i.e., the prior)
\end{itemize}
Further, for each hidden state $z_t$, corresponding observation $x_t$ is a realization of an emission process $\mathbf{B}=[b_j(x)]$ where $b_j(x)=p(x\mid z=q_j)$. We assume $b_j(x)$ follows a Gaussian mixture model (GMM) as defined in Equation~\ref{eqn:mixmod}. 


\begin{equation}
    p(x_t\mid z=q_j)= \sum_{l=1}^{k}c_{jl}\mathcal{N}(x_t\mid \mu_{jl},\Sigma_{jl})
\label{eqn:mixmod}
\end{equation}\
where $\sum_{l=1}^{k}c_{jl}=1, \forall j=\{1,\cdots,M\}$, $k$ is the number of Gaussian mixture components and $\mathcal{N}(x_t\mid \mu_{jl},\Sigma_{jl})$ denotes a Gaussian probability density with mean $\mu_{jl}$ and covariance $\Sigma_{jl}$ for state $j$ and mixture component $l$. The number of hidden states ($M$) and mixture components ($k$) are the two hyperparameters of the model which have to be provided apriori.  
    
Therefore, the joint probability density function of the observation $\mathbf{X}$ can be expressed as:
\begin{equation}
p(\mathbf{X})=p(z_1)\prod_{t=1}^{T-1}p(z_{t+1}\mid z_{t})\prod_{t=1}^{T}p(x_t\mid z_t)
\label{eqn:markov}
\end{equation}
The optimum parameters $[\mathbf{A},\mathbf{B}, \pi]$ that locally maximize the total observation likelihood (Equation~\ref{eqn:markov}) of observation $\mathbf{X}$, are estimated using an expectation-maximization algorithm, known as the Baum-Welch algorithm \cite{Rabiner1993}.
Furthermore, the probability of the system being in a given latent state, $z_{t}$  corresponding to $\mathbf{x_t}$ is computed using the Viterbi algorithm.

\subsubsection{Sojourn time distribution}
An inherent assumption of the HMM is that the number of time steps ($u$) spent in a given state $q_j$ (a.k.a sojourn time) is geometrically distributed (denoted by $d_j$) as shown below.
\begin{equation}
    d_j(u) = a_{jj}^u(1-a_{jj})\label{eqn:sojourn}
\end{equation}
However for some dynamical systems, the probability of a state change depends on the time spent in the current state. Therefore, geometrically distributed sojourn time fails to model such systems.
An alternate solution is to explicitly estimate the duration density $d(u)$, which results in a hidden \textit{semi}-Markov model (HSMM). In this study, we compare Gamma, Weibull and logarithmic distributions for sojourn density in addition to the default choice of geometric distribution. For each of these assumptions, the parameters of the HSMM model is estimated by maximizing the likelihood of the joint probability density function of $\mathbf{X}$, as shown in Equation~\ref{eqn:hsmm} \cite{Guedon2005}.

\begin{multline}
p(\mathbf{X})=p(z_1) d_{z_1}(u_1) \Bigl\{\prod_{t=1}^{T-1}  p(z_{t+1}\mid z_{t})d_{z_t}(u_t)\Bigr\} \\ p(z_{T}\mid z_{T-1})D_{z_T}(u_T) \prod_{t=1}^{T} p(x_t\mid z_t)
\label{eqn:hsmm}
\end{multline}
In the above equation, the survival function, $D_j(u)$, is used to represent the time spent in the last state since the system is not observed beyond time $T$. Using this survival function improves the parameter estimation and, provides a more accurate prediction of the last state visited.

\subsection{Long short-term memory}
Feed-forward neural network architectures are not explicitly designed to handle sequential data. A class of DL approaches, recurrent neural network (RNN), uses a feedback mechanism where the output from a previous time step is fed as an input to the current time step such that information from the past can propagate into future states. This feedback mechanism preserves the temporal correlation and makes it suitable to capture the temporal evolution of traffic parameters. However, RNNs are incapable of handling the long-term dependencies in temporal data due to the vanishing gradient problem \cite{hochreiter1998}. Long short-term memory (LSTM) \cite{hochreiter1997}, a type of RNN, consists of memory cells in its hidden layers and several gating mechanisms, which control information flow within a cell state (or, memory) to selectively preserve long-term information. 

\begin{figure}[!htb]
\centering
\includegraphics[width=0.9\columnwidth]{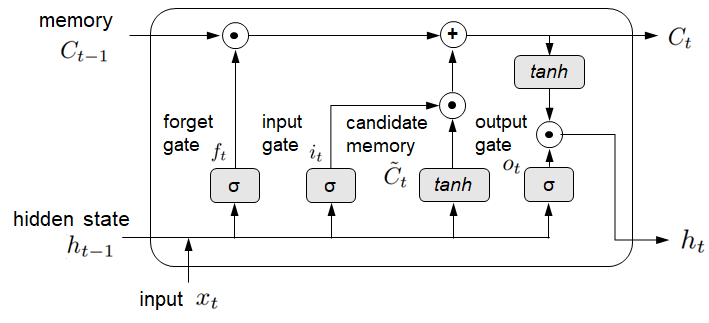}
\caption{Schematic diagram of LSTM module}\label{fig:lstm_schematic}
\end{figure}

The objective is to update the cell, $C_t$, over time using the input $x_t$ and the previous time step's hidden state, $h_{t-1}$. This process involves several key operations. First, a forget gate, $f_t$, selectively filters information from the past. Then, an input gate, $i_t$, regulates the amount of information from the candidate memory cell, $\Tilde{C}_t$, that should be incorporated into the current cell state, $C_t$. Finally, an output gate, $o_t$, governs the update of the hidden state, $h_t$. See Figure~\ref{fig:lstm_schematic}. The computations are represented as follows:

\begin{equation}\label{eqn:cellstate}
\begin{split}
    \Tilde{C}_t &= \tanh(W_{c}[{h}_{t-1},{x}_t] + b_{c})\\
    C_t &= f_t \odot C_{t-1} +i_t\odot \Tilde{C}_t\\
h_{t} &= o_{t}\odot \tanh(C_{t})\\
\end{split}
\end{equation}
The outputs from the forget gate, $f_{t}$, input gate, $i_{t}$, and output gate, $o_{t}$ are computed as shown below:
\begin{equation} \label{eqn:lstm_gates}
\begin{split}
f_t & = \sigma({W}_f[{h}_{t-1},{x}_t] + b_{f})\\
i_t & = \sigma({W}_i[{h}_{t-1},{x}_t] + b_{i})\\
o_t & = \sigma({W}_o[{h}_{t-1},{x}_t] + b_{o})\\
\end{split}
\end{equation}
Here, $\sigma$ and $\tanh$ represent non-linear activation functions, while $W_f$, $W_i$, $W_o$, and $W_{c}$ denote weight matrices corresponding to the forget gate, input gate, output gate, and candidate memory cell, respectively. Similarly, $b_f$, $b_i$, $b_o$, and $b_c$ represent the corresponding bias vectors.

\subsection{Modeling traffic data}
Vehicle arrival rates in traffic are commonly modeled as a Poisson process, assuming that vehicles arrive independently at a specific location or time according to Poisson distribution with a fixed average rate, $\lambda$. However, in real-world traffic scenarios, the average arrival rate $\lambda(t)$ fluctuates throughout the day, resulting in a non-homogeneous Poisson process. This means that the rate of vehicle arrivals vary over time, reflecting changes in traffic flow and congestion levels. 
This variation in the Poisson average rate captures the dynamic nature of traffic patterns, aligning with real-world observations of different traffic conditions at different times of the day, such as high flow periods during rush hours or lower flow periods during off-peak times. 

In our study, we analyze the data representing the fluctuations in vehicle counts, i.e., the changes in vehicle arrival at each time step. While the arrival rates in traffic are commonly modeled as a Poisson process, the observed fluctuation data follows a Skellam distribution, which arises from taking the difference between two Poisson random variables with parameters $\lambda(t)$ and $\lambda(t+\delta t)$, where $\delta t$ represents the time gap between consecutive observations. Throughout the day, the average rate of vehicle arrivals fluctuates, leading to variations in the parameters of the Skellam distribution. Consequently, the distribution of traffic fluctuation can be described as a mixture of Skellam distributions, with each component representing a specific average arrival rate.

In our approach, we model the temporal sequence of traffic fluctuations using a Hidden Markov Model (HMM). In other words, the HMM tries to categorize the outcome of the traffic fluctuations, which are assumed to be a random variable that is the mixture of Skellam distributions. More specifically, the HMM approximates the output space by employing a mixture of Gaussian distributions, allowing for effective modeling and inference of the underlying traffic patterns. However, since the HMM assumes a continuous distribution while the Skellam is a discrete distribution, the Skellam distribution is normalized using the mean and standard deviation of the Skellam distribution. 

Furthermore, it is worth noting that the independence of observations justifies the use of an HMM, as each observation in the traffic data is considered to be independent of previous observations. Moreover, despite the varying average trend in real traffic data throughout the day, the fluctuations tend to exhibit sporadic behavior, indicating little to no duration dependency. To account for this characteristic, multiple duration densities are employed to model the sojourn durations and find the distribution(s) that best fit the data.

\section{Methodology} 
The hidden Markov model (HMM) and long short-term Memory (LSTM) are two distinct models that generate latent space representations for modeling the distribution of an observed sequence. While they have structural similarities and divergences, they can be considered as special instances of a more comprehensive framework called the Generative Unified Model (GUM).

In the GUM framework, there exists a hidden or latent state which provides information about the observation. In the HMM, the hidden state $z$ follows Markovian dynamics, meaning that the current state $z_t$ depends only on the previous state $z_{t-1}$ and is conditionally independent of the observation $x$ (as represented by Equation~\ref{eqn:markov}). 
On the other hand, in the LSTM model, the latent variable $h_t$ is deterministic and is a function of the previous latent variable $h_{t-1}$ and the current observation $x_t$ (see Equation~\ref{eqn:cellstate}). The LSTM captures temporal dependencies in the sequence by updating its hidden state representation based on the previous state and current observation.

These structural dissimilarities between the HMM and LSTM result in the two models learning complementary feature representations of an observed sequence. This phenomenon has been highlighted in research conducted in the field of natural language processing (NLP) \cite{Achille2019,Liu2019,krakovna2016}. For instance, hybrid RNN-HMM architectures were explored in text sequence modeling \cite{krakovna2016}, where the HMM modeled the underlying statistical patterns in the input sequence, while the LSTM captured the temporal dependencies using its hidden state representation.

In the context of traffic prediction, we aim to leverage this complementary feature learning phenomenon by combining HMM and LSTM in two proposed architectures. 
By combining these two models, we can effectively capture both the statistical patterns and the temporal dynamics of the traffic data, leading to enhanced prediction accuracy. 
This stands in contrast to the baseline model, which is a simple LSTM model (See Figure~\ref{fig:architecture}(a)) that solely relies on the temporal history of $x$ to predict its future value. Additionally, unlike multi-regime models that employ separate prediction models for different states, the hybrid models train a single prediction model that captures the system's evolution within the latent space.

\subsection{Model architectures}
In this study, two architectures of a hybrid HMM-LSTM model are considered: the sequential hybrid (S-Hybrid) and the concatenated hybrid (C-Hybrid).

\subsubsection{Sequential hybrid}
In the sequential hybrid model (S-Hybrid), the first step is to train an HMM on the input sequence $\mathbf{X}$ to learn the probability of the system being in each hidden state $q \in Q$ at time $t$. The HMM captures the time-evolution of state probabilities based on the observed sequence. These HMM features, representing the probabilities of being in different hidden states, are then used as input to train the LSTM to learn the temporal dependencies in the sequence.

In the sequential Hybrid Model (S-Hybrid), the initial step involves training an HMM on the input sequence $\mathbf{X}$ to estimate the likelihood of the system being in each hidden state $q \in Q$ at time $t$. The HMM effectively captures the dynamic changes in state probabilities based on the observed sequence. These HMM features, which represent the probabilities associated with different hidden states, are subsequently utilized as input for training the LSTM network to learn the temporal dependencies within the sequence. 
The latent outputs ($h$) from the LSTM are processed through a series of dense layers to generate the final prediction. This S-Hybrid approach effectively combines the probability information obtained from the HMM with the LSTM's capabilities, enhancing the model's ability to anticipate state transitions. As a result, it is expected that this modeling approach can potentially lead to improved prediction performance by leveraging the complementary strengths of the HMM and LSTM. See Figure~\ref{fig:architecture}(b) for the architecture.

\subsubsection{Concatenated hybrid}

\begin{figure*}[!htb]
\centering
\includegraphics[width=0.8\textwidth]{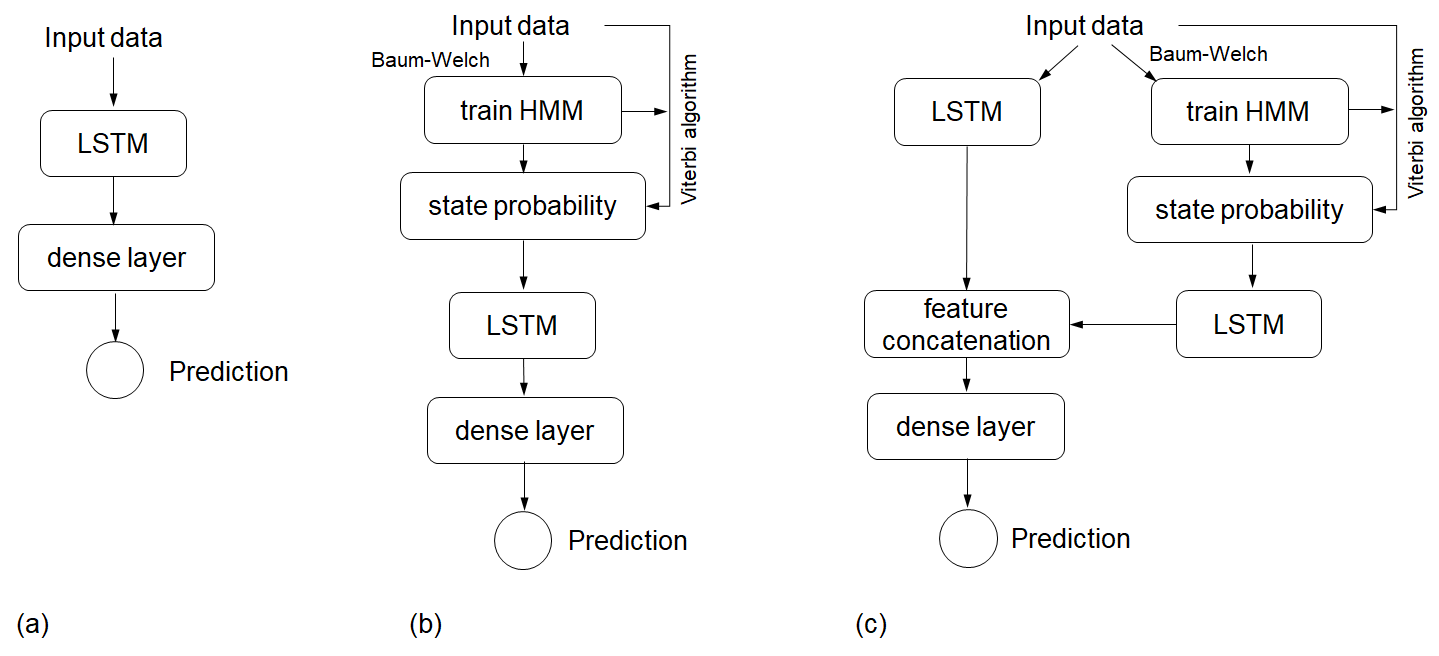}
\caption{DL architectures considered (a)Stacked LSTM; (b) Sequential Hybrid; (c) Concatenated Hybrid}\label{fig:architecture}
\end{figure*}

In the concatenated hybrid model (C-Hybrid), the latent outputs from two distinct LSTM networks are combined. One LSTM network is trained on the input sequence $X$, while the other LSTM network is trained on the sequence of hidden states obtained from the HMM. These two sets of latent outputs, representing the learned temporal dependencies from both the input sequence and the HMM state sequence, are concatenated together. The concatenated features are then fed into a series of densely connected layers to generate the final prediction. By integrating the latent outputs from the LSTM networks trained on different sequences, the model potentially captures the complementary information and leverages it to enhance prediction accuracy. This approach allows the model to benefit from both the statistical patterns captured by the HMM and the temporal dynamics captured by the LSTM. See Figure~\ref{fig:architecture}(c) for the architecture.

\subsection{Model training}

The baseline deep learning (DL) model employed in this study consists of a stacked LSTM architecture. Our model consists of three LSTM layers with 20, 20 and 10 units, followed by four dense layers with 10, 10, 6 and 2 units respectively, with LeakyReLU activation function \cite{maas2013rectifier} for dense layers respectively. 
Additionally, a statistical benchmark model called the HMM-based regime-switching autoregressive model (AR-HMM) \cite{kim2008} was chosen for comparison. The AR-HMM is a widely recognized model frequently used in traffic data analysis literature.

The hyperparameters of the proposed architectures were selected to ensure that the number of trainable parameters is comparable for each DL model. Specifically, the same architecture is utilized for the S-Hybrid model as the LSTM model, with the only difference being the number of feature channels in S-Hybrid, which is set equal to the number of hidden HMM states considered. 
In the case of the C-hybrid model, it consists of two branches. Each branch comprises two LSTM layers with 20 and 10 units, respectively. The feature outputs from these branches are merged, and then passed through four dense layers with 10, 6, 6, and 1 units, respectively with LeakyReLU activation.

To ensure the generalizability of the models, the models are trained, validated, and tested on three separate sets. The dataset is divided into three parts: 60\% for model training, 15\% for validation, and 25\% for testing. This division allows for assessing the model's performance on unseen data and helps prevent overfitting. To address overfitting, the model parameters are tuned throughout the training process based on their performance on the validation set. 
The models are trained to minimize the mean squared error (MSE) loss function for sufficiently large number of epochs until the validation loss starts to increase. The model with the lowest validation error is selected as the final model.
We use Adadelta \cite{adadelta} as the optimizer with the learning rate of 0.20, $\rho$ value of 0.95, and epsilon of 1e-7 to train the models.

\section{Data}
The performance evaluation of the proposed models is conducted on a dataset obtained from the California Department of Transportation's Performance Measurement System (PeMS). This dataset is widely used for traffic data modeling. The traffic data, including flow, occupancy, and speed, is collected from vehicle detector stations (VDS) located along freeways and ramps. The raw data is sampled at 30-second intervals and then aggregated at 5-minute intervals.
Flow data for one year (i.e., 104942 samples) from VDS 1114805 on California Interstate 05 NB in District 11 were used for training and testing of the models. The performance of the models are evaluated on 25\% of the dataset that was never used in training. 

\begin{figure}[!htb]
\centering
\includegraphics[width=1\columnwidth]{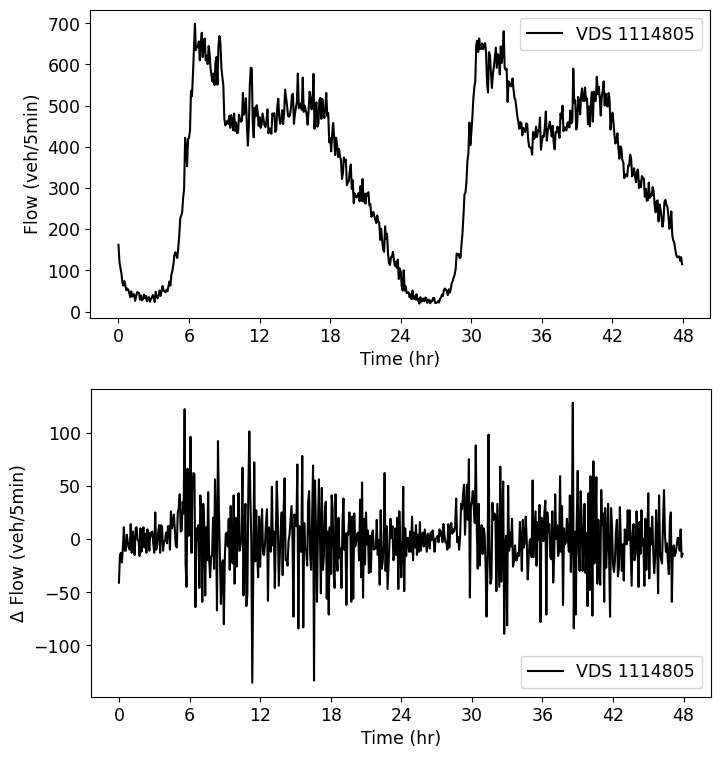}
\caption{(a)Traffic flow; (b) detrended flow or fluctuation for 24-hr period}\label{fig:detrend}
\end{figure}

In this study, the models are employed to predict fluctuations in traffic flow, specifically the change in flow between successive time steps ($\Delta~\text{Flow}$), instead of directly predicting the absolute flow values at a detector location. It is important to note that modeling the first-order difference, i.e., the flow fluctuations, can be advantageous for capturing short-term dynamics in the time series, whereas, modeling the actual time series can be more suitable for capturing long-term trends and patterns \cite{li2014}. Figure~\ref{fig:detrend} shows the detrended time series ($\Delta~\text{Flow}$) corresponding to the flow observed at the detector during a 48-hour cycle. 

\section{Results}
In this study, three DL models -- 1) a stacked LSTM model trained on flow fluctuations, 2) a stacked LSTM model trained on probabilities of hidden state transition (or, S-Hybrid) and 3) merged LSTM model that takes both flow fluctuations and probabilities of hidden state transition as inputs, are considered for prediction tasks. 

We evaluate the model performances for single-step prediction horizons, by comparing the prediction mean with the corresponding true values using three metrics: root mean squared error (RMSE), mean absolute percentage error (MAPE) and $R^2$ as defined below. 

\begin{equation} 
\mathrm{RMSE}={\sqrt{\dfrac{1}{n}\sum_{i=1}^{N} \left [y_{i}-\hat {y_{i}}\right]^{2}} }\label{eqn:rmse}
\end{equation}

\begin{equation} 
\mathrm{MAPE}=\dfrac{1}{N}\sum_{i=1}^{N} \lvert{ \dfrac{y_{i}-\hat {y_{i}}}{y_{i}} }\rvert \label{eqn:mape}
\end{equation}

\begin{equation} 
\mathrm{R^2}=1-\dfrac{\sum_{i=1}^{N} \left [y_{i}-\hat {y_{i}}\right]^{2}}{\sum_{i=1}^{N} \left [y_{i}-\bar {y_{i}}\right]^{2}} \label{eqn:mae}
\end{equation}

where $y_i$ represents the 'ground truth' or true value of the observation $i$, $\hat{y_i}$ is the predicted value of $y_i$ for $i=1,2,\dots T$. 

In the `Methodology' section, the hybrid models are described as utilizing HMM features derived from the input data, specifically the flow fluctuations ($\Delta~\text{Flow}$), to perform the prediction task. To enable this, HMMs are trained on the input data with different configurations. In this study, HMMs are trained assuming 3 and 5 latent states, and a Gaussian mixture model is used with either 1 or 2 mixture components.

To characterize the duration densities of the states, various distributions such as geometric, logarithmic, gamma, and Weibull distributions are assessed. For each case studied, the AIC (Akaike Information Criterion) and BIC (Bayesian Information Criterion) are computed to select the models based on a trade-off between model fit and complexity. The results are presented in Table~\ref{tab:perf_test}. It is worth noting that the AIC and BIC values for models with 1 and 2 Gaussian mixture components are consistently similar. Hence, for the sake of brevity and reduced model complexity, results obtained with 1 Gaussian component are reported.
Across all the evaluated distributions, the AIC and BIC values indicate that a model with 3 latent states is preferable. 
This configuration provides a good balance between model fit and complexity. On the other hand, using 5 latent states introduces more complexity, without substantial improvement in model performance, resulting in higher AIC and BIC values. 
\footnote{The computations for these evaluations are conducted using the Hidden Markov Model package \cite{amini2022}.
}

\begin{table*}[h!]
\caption{Comparison of AIC and BIC values for different HMM configurations}\label{tab:perf_test}	
\begin{center}
\begin{tabular}{llllll}
\hline
\multirow{2}{*}{Metric} & \multirow{2}{*}{State} & \multicolumn{4}{c}{Sojourn distribution} \\ \cline{3-6} 
 &  & Geometric & Logarithmic & Gamma & Weibull \\ \hline
AIC & 3 & 196652.6 & 192302.1 & 216200.9 & 213386.6 \\ \cline{2-6} 
 & 5 & 216445.4 & 201293.3 & 242818.6 & 233851.3 \\ \hline
BIC & 3 & 196819.5 & 192496.8 & 216395.6 & 213581.4 \\ \cline{2-6} 
 & 5 & 216816.3 & 201692 & 243217.3 & 234250 \\ \hline
\end{tabular}
\end{center}
\end{table*}

\begin{figure*}[!htb]
\centering
\includegraphics[width=0.95\textwidth]{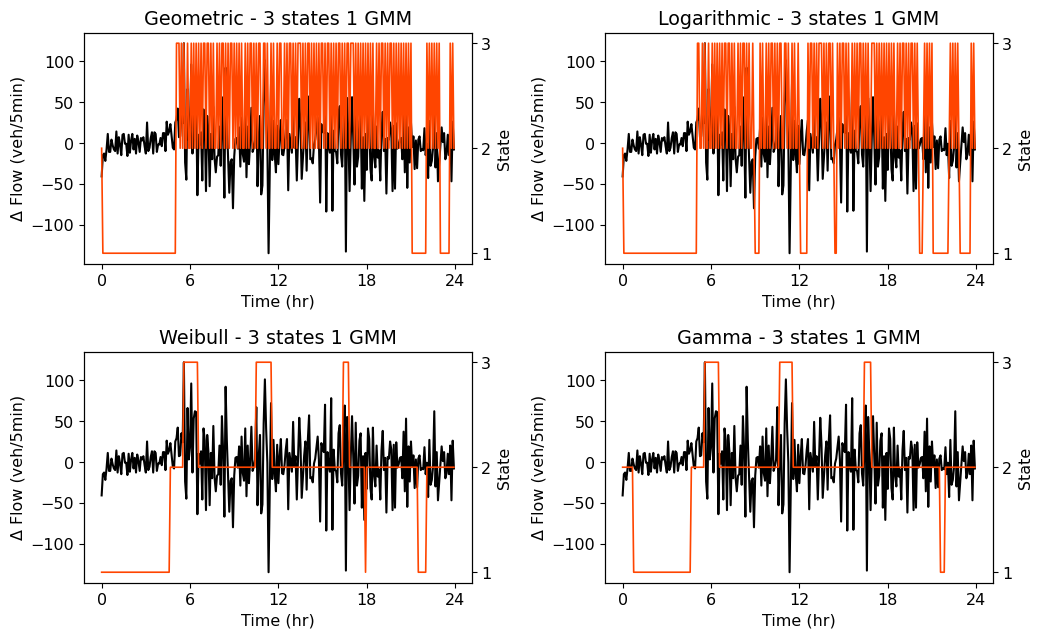}
\caption{Identified hidden states by different sojourn densities with 3 latent states and 1 Gaussian mixture component}\label{fig:state_3}
\end{figure*}

\begin{figure*}[h!]
\centering
\includegraphics[width=0.95\textwidth]{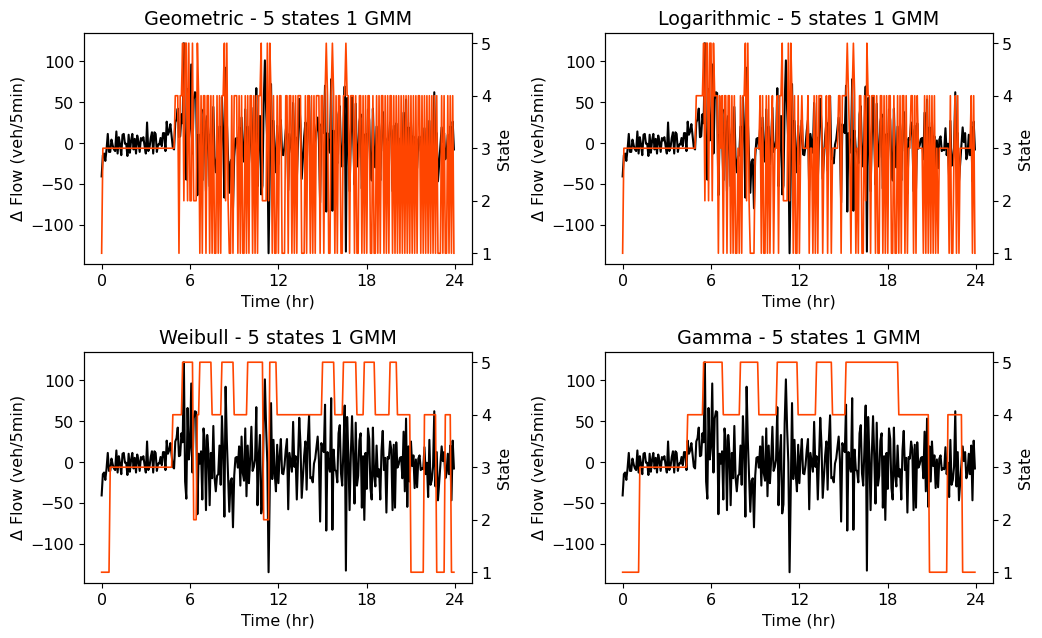}
\caption{Identified hidden states by different sojourn densities with 5 latent states and 1 Gaussian mixture component}\label{fig:state_5}
\end{figure*}

Using the trained hidden (semi) Markov models, the most likely states for each time instant is identified as shown in Figures~\ref{fig:state_3} and \ref{fig:state_5}. As can be seen, the system dynamically transitions between the hidden states within the day. However, due to different sojourn densities, the identified states and their duration are quite different. 
In case of geometric and logarithmic sojourn density, the system is observed to exhibit abrupt state changes as flow increases and congestion builds on the roadway. Therefore, traffic is observed to be unstable during congestion, as highlighted by \cite{disbro1989}. 
Conversely, the system is observed to remain in each state for an increased time duration when Gamma and Weibull sojourn distributions are assumed. Therefore, abrupt fluctuations in flow during congestion are not efficiently captured in these two cases. 

The performance of the S-Hybrid and C-Hybrid DL models for different configurations of hidden (semi) Markov models are compared in Tables~\ref{tab:perf_seq} and \ref{tab:perf_merged} respectively. Upon analyzing the results, it is evident that models utilizing geometric and logarithmic sojourn densities demonstrate superior performance compared to models using Gamma and Weibull sojourn densities. This observation is consistent with the characteristics of the observed state transitions in the time-series data as depicted in Figures~\ref{fig:state_3} and \ref{fig:state_5}.

\begin{table*}[!ht]
\caption{Performance comparison of S-Hybrid DL models with different sojourn densities}\label{tab:perf_seq}	
\begin{center}
\begin{tabular}{llllll}
\hline
\multirow{2}{*}{Metric} & \multirow{2}{*}{State} & \multicolumn{4}{c}{Sojourn distribution} \\ \cline{3-6} 
 &  & Geometric & Logarithmic & Gamma & Weibull \\ \hline
\multirow{2}{*}{RMSE} & 3 & 0.6163 & 0.6229 & 0.9486 & 0.9493 \\ \cline{2-6} 
 & 5 & 0.5635 & 0.5326 & 0.9457 & 0.9447 \\ \hline
\multirow{2}{*}{$R^2$} & 3 & 0.5882 & 0.5794 & 0.0246 & 0.0230 \\ \cline{2-6} 
 & 5 & 0.6557 & 0.6925 & 0.0305 & 0.0323 \\ \hline
\multirow{2}{*}{MAPE} & 3 & 168.8370 & 155.0041 & 52.4545 & 48.9196 \\ \cline{2-6} 
 & 5 & 167.3981 & 142.0781 & 47.6088 & 48.2944 \\ \hline
\end{tabular}
\end{center}
\end{table*}

\begin{table*}[h]
\caption{Performance comparison of C-Hybrid DL models with different sojourn densities}\label{tab:perf_merged}	
\begin{center}
\begin{tabular}{llllll}
\hline
\multirow{2}{*}{Metric} & \multirow{2}{*}{State} & \multicolumn{4}{c}{Sojourn distribution} \\ \cline{3-6} 
 &  & Geometric & Logarithmic & Gamma & Weibull \\ \hline
\multirow{2}{*}{RMSE} & 3 & 0.4631 & 0.4898 & 0.8106 & 0.8167 \\ \cline{2-6} 
 & 5 & 0.4245 & 0.4203 & 0.8055 & 0.8019 \\ \hline
\multirow{2}{*}{$R^2$} & 3 & 0.7675 & 0.7400 & 0.2876 & 0.2769 \\ \cline{2-6} 
 & 5 & 0.8046 & 0.8085 & 0.2967 & 0.3029 \\ \hline
\multirow{2}{*}{MAPE} & 3 & 150.3153 & 148.6007 & 146.0438 & 147.1233 \\ \cline{2-6} 
 & 5 & 138.7888 & 116.3367 & 143.3267 & 145.9820 \\ \hline
\end{tabular}
\end{center}
\end{table*}

The geometric distribution assumes a constant probability of transitioning from one state to another, regardless of the duration already spent in the current state. 
On the other hand, the logarithmic distribution considers the time already spent in a state, allowing for a wider range of durations and better adaptation to the observed patterns in the traffic data. This flexibility in capturing varying durations enables the logarithmic sojourn density to more effectively model the complex dynamics of traffic behavior, leading to slightly better performance compared to the geometric sojourn density. This aligns well with the assumption of vehicle arrival being modeled as a Poisson process and the (near) independence assumption.

In this study, a logarithmic sojourn density with 5 states and 1 Gaussian emission is found to marginally outperform the geometric sojourn density in terms of prediction accuracy for both the S-Hybrid and C-Hybrid DL models.
The optimized parameters for the fitted HMM model with logarithmic sojourn density, 5 states, and 1 Gaussian emission distribution are denoted by Equation~\ref{eqn:hmm_para}. 

\begin{equation} \label{eqn:hmm_para}
\begin{split}
A & = \begin{pmatrix}
0.000 & 0.0161 & 0.258 & 0.714 & .012\\
0.109 & 0.000 & 0.009 & 0.046 & 0.836\\
0.441 & 0.002 & 0.000 & 0.548 & 0.009\\
0.768 & 0.008 & 0.144 & 0.000 & 0.081\\
0.038 & 0.572 & 0.029 & 0.361 & 0.000\\
\end{pmatrix}\\
\pi & = \begin{bmatrix}
1 & 0 & 0 & 0 & 0\\
\end{bmatrix}\\
p & = \begin{bmatrix}
0.307 & 0.455 & 0.934 & 0.420 & 0.245\\
\end{bmatrix}\\
s & = \begin{bmatrix}
1 & 1 & 1 & 1 & 1\\
\end{bmatrix}\\
\mu & = \begin{bmatrix}
-0.8594 & -0.7903 & -0.0098 & 0.8144 & -0.8524\\
\end{bmatrix}\\
\sigma & = \begin{bmatrix}
0.3430 & 6.1145 & 0.1494 & 0.4009 & 2.3412\\
\end{bmatrix}\\
\end{split}
\end{equation}\\
where $A$ represents the hidden state-transition matrix, $\pi$ corresponds to the initial state probabilities, $p$ and $s$ denote the scale and shift parameters of the sojourn density, and $\mu$ and $\sigma$ represent the parameters of the Gaussian emission function.

\begin{figure*}[!]
\centering
\includegraphics[width=0.95\textwidth]{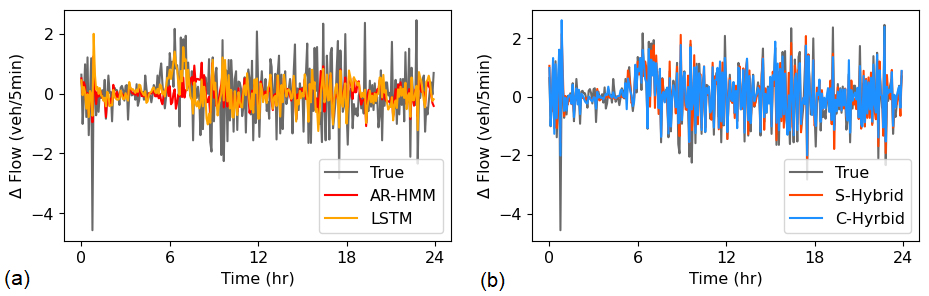}
\caption{Performance comparison of (a) LSTM and AR-HMM models, (b) Hybrid models in 24-hr period}\label{fig:perf}
\end{figure*}

The proposed DL models with logarithmic sojourn density with 5 states and 1 Gaussian emission are compared with two baseline models -- a stacked LSTM and HMM-based regime-switching autoregressive model (AR-HMM) with lags 1, 10 and 20.  
As observed from Table~\ref{tab:perf}, the hybrid models perform significantly better than the baseline models, with C-Hybrid outperforming S-Hybrid. 
Figure~\ref{fig:perf} demonstrates the prediction performance for a 24-hour period. It is evident from the figure that AR-HMM and LSTM follow similar trends, while the hybrid models perform significantly better to capture the abrupt flow changes. 
However, in the free-flow regime (2 to 6 hrs), S-Hybrid fails to capture the trend. This is due to the fact that the system predominantly remains in state 3 during free-flow and hence, the input to the LSTM i.e., the hidden state probabilities over time does not change appreciably. Therefore, the model generates a near constant output. To the contrary, the predictions of C-Hybrid performs comparatively better than S-Hybrid to capture the low-fluctuations. 

\begin{table}[!ht]
\caption{Performance comparison of models}\label{tab:perf}	
\begin{center}
\begin{tabular}{lllr}
\hline
Model name & RMSE & $R^2$ & MAPE \\ \hline
LSTM       &  0.8235    &  0.2648 & 164.9522   \\
1-lag AR-HMM     &  0.9083   &  0.1046 & 98.0325    \\ 
10-lag AR-HMM     &  0.8843   &  0.1522 & 109.7145   \\ 
20-lag AR-HMM     &  0.8766   &   0.1669 & 125.6203    \\
S-Hybrid   &  0.5326   &  0.6925 & 142.0781   \\ 
C-Hybrid   &  0.4203    & 0.8085 & 116.3367 \\ \hline
\end{tabular}
\end{center}
\end{table}

Further, we compare the feature-space representations of the penultimate layers of the models to identify specific patterns that enhance prediction capabilities of hidden Markov-LSTM models. We use $t$-Stochastic Neighbor Embedding, a non-linear technique for dimension reduction, to reduce the high dimensional feature output to two dimensions \cite{van2008}.  
Flow fluctuation data were labeled into four traffic regimes: 1) low flow (0 to 6 hr), 2) increasing flow (6 to 8 hr), 3) high flow (8 - 18 hr) or congestion and 4) decreasing flow (18 to 24 hr) which are suitably color-coded as shown below.

\begin{figure}[!]
\centering
\includegraphics[width=1\linewidth]{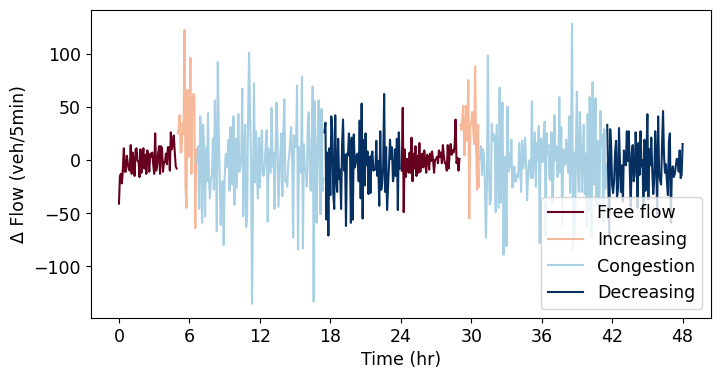}
\caption{Flow and fluctuation segmented based on regimes}
\label{fig:feature}
\end{figure}

Figure~\ref{fig:feature} illustrates the learned feature space for the four different regimes obtained using the stacked LSTM, S-Hybrid, and C-Hybrid models. When considering the LSTM model, it is evident that different traffic states overlap, making it challenging to distinguish between them. However, with the incorporation of HMM features into the model, we observe clear separations in the feature space for the hybrid models. Notably, the C-Hybrid model exhibits superior separability for low flow traffic states (as shown in Figure~\ref{fig:feature}), which likely contributes to its superior performance. Table~\ref{tab:tsne} presents a comparison of the variances of outputs belonging to specific traffic regimes in the 6-dimensional feature space for each model. It is worth noting that the LSTM model demonstrates high dispersion of outputs within the feature space for data from the same traffic states, along with significant overlap between data from multiple regimes. In contrast, the hybrid models incorporating HMM features effectively localize features in the space, resulting in enhanced performance.

\begin{figure*}[!]
\centering
\includegraphics[width=0.9\linewidth]{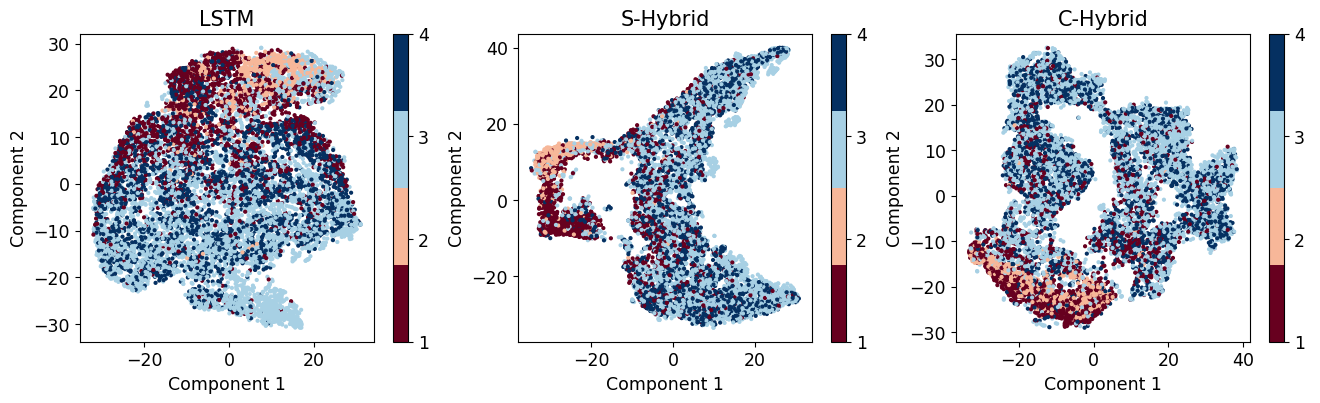}
\caption{Feature space visualizations of layer outputs of LSTM, S-Hybrid and C-Hybrid models}
\label{fig:feature}
\end{figure*}

\begin{table}[]
\caption{Variance of 6-dimensional outputs of models for different traffic flow regimes }\label{tab:tsne}
\begin{center}
\begin{tabular}{lccc}
\hline
Regime     & LSTM & S-Hybrid & C-Hyrbid \\ \hline
Low        & 0.1709   & 0.0470    & 0.0724\\ 
Increasing &  0.1458   & 0.0294    & 0.0783\\
High       & 0.2890   &  0.1284    & 0.0862\\
Decreasing & 0.2399   & 0.1046    & 0.0769\\ \hline
\end{tabular}
\end{center}
\end{table}



\section{Conclusions}
Hidden Markov model (HMM) and recurrent neural networks like Long short-term memory (LSTM) are capable of modeling an observation sequence from a set of latent (hidden) state variables. The latent variables in LSTM are determined in a deterministic manner from the current observation and the previous latent variable, while, in HMM, the set of latent variables is a Markov chain. 
Recent research highlights the structural similarity between LSTM and HMM, and their capability to learn complementary features from input data. Therefore, appropriate hybridization of these models could lead to a better modeling of the data compared to the individual models. 

Our study adopts a hybrid approach combining a HMM and LSTM to model the temporal sequence of traffic fluctuations. Specifically, the HMM allows to capture the underlying patterns and state changes in traffic dynamics that typically are assumed to have a Poisson distribution. The HMM's capability to model the short-term fluctuations in traffic, often resembling a Skellam distribution, enables us to accurately characterize the variability in traffic behavior. Additionally, we incorporate the LSTM, which retains long-term temporal correlations, allowing us to capture complex dynamic patterns and non-stationarity within the traffic data. 
As shown in this study, hybrid models that jointly use HMM and LSTM to perform the task of traffic flow prediction outperform in terms of prediction accuracy the LSTM and auto-regressive HMM regime switching models in capturing chaotic behavior in traffic data by learning complementary features. 
The testing of the models on loop detector data shows that the the LSTM and AR-HMM models result in an RMSE of 0.8235 and 0.8766, respectively, while the S-Hybrid and C-Hybrid models result in an RMSE of 0.5326 and 0.4203, respectively, which corresponds to an approximately 31-52\% improvement in performance. Transferability of hybrid hidden Markov-LSTM models to predict traffic on out-of-distribution datasets will be explored in future studies.

\bibliography{manuscript.bib}  
\end{document}